\documentclass[sigconf]{acmart}
\AtBeginDocument{%
  \providecommand\BibTeX{{%
    \normalfont B\kern-0.5em{\scshape i\kern-0.25em b}\kern-0.8em\TeX}}}

 \setcopyright{acmcopyright}
 \copyrightyear{2023}
 \acmYear{2023}

\acmConference[AIMLSystems 2023]{ The Third International Conference on Artificial Intelligence and Machine Learning Systems}{October 25--28, 2023}{Bangalore, India}
\usepackage{algorithm}
\usepackage{algorithmic}
\usepackage{enumitem}
\begin{document}

\title[FENCE: Real-Time Multiple ID Detection at Scale In Fantasy Sports]{FENCE: Fairplay Ensuring Network Chain Entity for Real-Time Multiple ID Detection at Scale In Fantasy Sports}



\author{Akriti Upreti}
\email{akriti@dream11.com}
\authornote{All authors contributed equally to this research.}
\email{akritiupreti@utexas.edu}
\affiliation{%
  \institution{Dream11}
  \city{Mumbai}
  \country{India}
}
\author{Kartavya Kothari}
\authornotemark[1]
\email{kartavya.kothari@dream11.com}
\email{kartavya@cse.iitb.ac.in}
\affiliation{%
  \institution{Dream11}
  \city{Mumbai}
  \country{India}
}
\author{Utkarsh Thukral}
\authornotemark[1]
\email{utkarsh.thukral@dream11.com}
\affiliation{%
  \institution{Dream11}
  \city{Mumbai}
  \country{India}
}
\author{Vishal Verma}
\authornotemark[1]
\email{vishal.verma@dream11.com}
\affiliation{%
  \institution{Dream11}
  \city{Mumbai}
  \country{India}
}

\begin{abstract}

Dream11 takes pride in being a unique platform that enables over 190 million fantasy sports users to demonstrate their skills and connect deeper with their favorite sports. While managing such a scale, one issue we are faced with is duplicate/multiple account creation in the system \cite{amit_interview}. This is done by some users with the intent of abusing the platform, typically for bonus offers. The challenge is to detect these multiple accounts before it is too late. We propose a graph-based solution to solve this problem in which we first predict edges/associations between users. Using the edge information we highlight clusters of colluding multiple accounts. In this paper, we talk about our distributed ML system which is deployed to serve and support the inferences from our detection models. The challenge is to do this in real-time in order to take corrective actions. A core part of this setup also involves human-in-the-loop components for validation, feedback, and ground-truth labeling.

\end{abstract}



\maketitle

\section{Introduction}
``\verb |FENCE|'' (short for FairPlay Ensuring Network Chain Entity) is Dream11’s in-house multiple ID detection system. It is powered by a graph database that is responsible for processing and maintaining all models and heuristics so that FairPlay Violations are detected timely and efficiently.

The ``\verb |Fair|'' Play Policy at Dream11 defines a number of rules that users must adhere to in order to maintain a fair playing ground for all. Defying these rules constitutes a Fair Play Violation (“FPV”). A major kind of Fair Play Violation is the creation of Multiple ID (MI) a.k.a. duplicate accounts by a user on the platform in order to abuse referral or promotional cash bonus schemes which causes business heavy financial losses. These accounts are generally one-time use accounts that stay abandoned on the platform once their purpose is met. 

With ``\verb |FENCE| ''  we aim to identify and map the MIs to a single account. The key challenges are as follows. 
Firstly, we have limited information about a new user at the time of account creation which is detrimental to both detection recall and precision.
Secondly, delayed detection is as good as no detection. This is because delay allows the time for misuse of such accounts. It is essential to detect these accounts as soon as possible so as to circumvent their misuse hence the need for a real-time service.
Thirdly, even if at a delay, these accounts still need to be detected so that business metrics can be reported accurately. Additionally, it is important to collect this data which later can be valuable feedback to the real-time systems enhancing overall detection performance.
Lastly, in a fantasy sports platform like Dream11, there can be multiple matches running in parallel with varying levels of popularity. This implies that there can be high variance in the scale of users visiting and registering on the platform. Hence the system should deliver at all times.

To solve the above problems we propose FENCE. FENCE is a graph-based Machine Learning System that combines the worlds of supervised and unsupervised detection as well as real-time (transactional) and delayed(batch) processing. In doing so we are able to use the data as and when it becomes available. We use heuristics as edges between users as nodes to create a graph. In real time we use traversal queries to detect connected components. Using a threshold based on confidence level a bulk of these connected components are sent to a back-end service for real-time blocking. As more data becomes available we enrich the graph using probabilistic edges that are outputted from our Edge Prediction module. At the end of every day, we perform the batch processing. We collapse the graph into an edge list which is on the scale of billions. On this edge list, we run the Alternating Algorithm to determine revised connected components. Fresh highlights from this activity are sent for manual processing to our risk operation teams.

\section{Related Work}

Different methods have been used to identify fake accounts and Sybil accounts in online social networks, including user behavior analysis, graph theory application, machine learning techniques application, and robust system design.

The two main categories of conventional techniques for spotting multiple-ID accounts are feature-based techniques\cite{cao2014uncovering}\cite{egele2015towards}\cite{mondal2012defending} and structure-based techniques\cite{boshmaf2016integro}\cite{danezis2009sybilinfer}\cite{wang2017sybilscar}\cite{liang2021unveiling}. Machine learning approaches are commonly used in feature-based algorithms, which treat the identification of multiple-ID accounts as a binary classification problem. They start by collecting numerous elements from the content, behaviors (such as the sort of contest played, and the opponent's competence), and registration data (such as IP address and agent) of the account. Then, a supervised classifier is trained using these features using a labeled training set that includes both fraudulent and real accounts. Then, the trained classifier is used to spot bogus accounts.

On the other hand, structure-based methods take advantage of the structure of the social graph to spot multiple-ID accounts. They operate under the premise that if an account is linked to other fraudulent accounts, it is probably also fraudulent. These strategies frequently use graph-based machine learning techniques to examine the relationships in the social network in order to identify bogus accounts. Despite the fact that system-design strategies might be useful in preventing abuse, they are frequently hard to apply in large-scale networks that were initially created with the goal of maximizing growth and user engagement before abuse became a significant issue.

A detection method that bears the closest resemblance to our own research has recently been proposed by \cite{liang2021unveiling}. They similarly use a graph to perform community detection, comparing the efficacy of unsupervised detection over supervised detection methods. Our approach uses the above as a sub-module of a complex system that combines supervised and unsupervised approaches. Furthermore, we enhance our contributions by providing detailed insights into each component of the technology stack, particularly addressing the challenges that arise when implementing in real-time as the scale of the system grows which is not covered exhaustively in past work.

\section{Problem Formulation}

The problem statement is to detect multiple IDs on the platform that are directly or indirectly linked to only one user behind the screen. This is to ensure users’ compliance with FairPlay provisions of the platforms as touched upon above.  The number of duplicates can vary. It could be a single user controlling 2, 3, or even up to a few thousand accounts on the platform. 

We propose below a conceptual two-step formulation of the problem:

\subsection{Step 1: Edge prediction}

We predict an ‘edge’ between two given users on the platform. We define ‘edge’ as a link or a relation, broadly a form of the potential connection between two users based on certain characteristics \cite{sadowski2014fraud}. The idea is to vet each incoming user (say, a newly registered user at time ‘t’) by determining if there is any possible relationship between it and those users already registered on the platform before time ‘t’. Technically, the step will produce an output edge list ‘E’ containing all possible edges between a newly registered user and existing users on the platform. This edge list will then be used to build a graph that will be used in step 2.

\subsection{Step 2: Connected components creation}

The output from step 1 is used for determining the connected components \cite{cc}. We define a connected component as a connected sub-graph that is not part of any larger connected sub-graph.  Find  \(C \subseteq  {\left\{u_1,u_2,u_3,u_4,\cdots, u_n \right \}}\) is a connected component consisting of users  $u_1\: to\: u_n$ such that there exists an $x$ degree connection between users in $C$ where $x \in \left\{ Z \right\}^+ $  is a positive integer.

\section{Preliminaries}

\subsection{Real-time and batch detection flows}

The task of highlighting these multiple IDs is time-critical. Delayed detection of multiple accounts in the system can have business consequences. While the focus is on highlighting MI users immediately at registration via real-time flow, in certain cases we gain more confidence based on user interactions after they spend some time on the platform. Due to this difference in processing, MI(multiple ID) users can be highlighted in a real-time or delayed batch fashion. We use a combination of Spark's streaming and batch services to achieve this.

\subsection{Manual and automated actioning flows}

These connected components of colluding users are highlighted with scores ranging between 0 and 1. A high score greater than 0.95 implies very high chance of multiple ID collusion. Such highlights are sent to be processed via the automated flows. Highlights which have a score less than 0.95 or that come from the batch processes are acted upon through manual flows. The manual flow involves a "human-in-the-loop" who validates all highlights manually before taking any action. The feedback that we receive from manual reviews is tracked and maintained in our data warehouse tables. This data acts as a reliable ground truth to update and improve existing models thus reducing false positives over time.

\begin{figure*}[]
  \centering
  \includegraphics[width=\textwidth]{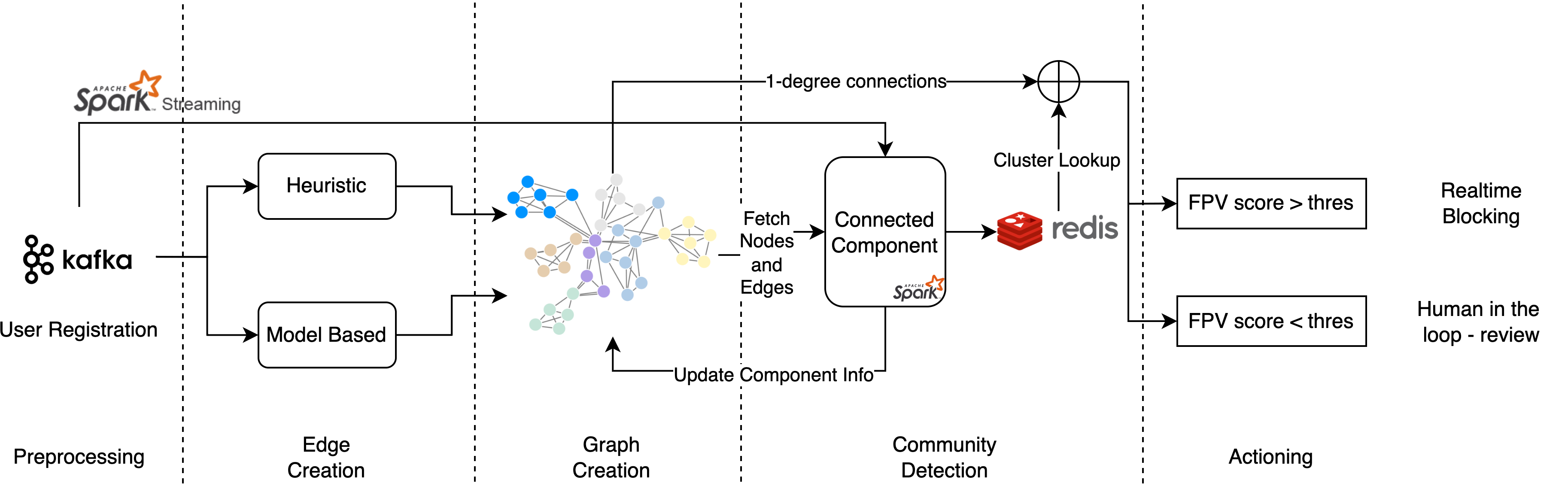}
  \caption{FENCE - ML system design}
  \label{fig:sys_des}
  \Description{End-to-end ML system design}
\end{figure*}

\section{ML System Design}
\subsection{Overview}
FENCE's objective is to identify fraudulent accounts both during their registration and after registration by employing a combination of unsupervised (real-time) and supervised (batch) methods. Figure \ref{fig:sys_des} provides an overview of FENCE, which comprises of 5 key components: \textit {1. Pre-processing of user-registration attributes},\textit{ 2. Edge creation, 3. Graph creation, 4. Multiple ID account detection, and 5. User actioning.}

In this section, we will discuss the overall design of FENCE's ML system, challenges faced at the production level with respect to real-time blocking, and the approaches taken to overcome these challenges.
\subsection{Pre-processing of user registration attributes}
\label{sec:Pre-Processing}
When a user registers on our app, several registration attributes are gathered related to the account. These attributes are subsequently processed to extract features that represent associations between two or more users. For example in the case of IP address, we receive a list of IP addresses against each user. We process this to extract the latest IP. Similarly, attributes like date of birth are converted to encrypted formats. Other device attributes are also extracted and cleaned. Since this is a critical system deployed currently in production we have to refrain from providing our complete attribute list. All these user-related attributes are then used in the edge creation step as detailed below.

\subsection{Edge creation}
\begin{figure}[h]
  \centering
  \includegraphics[width=0.8\linewidth]{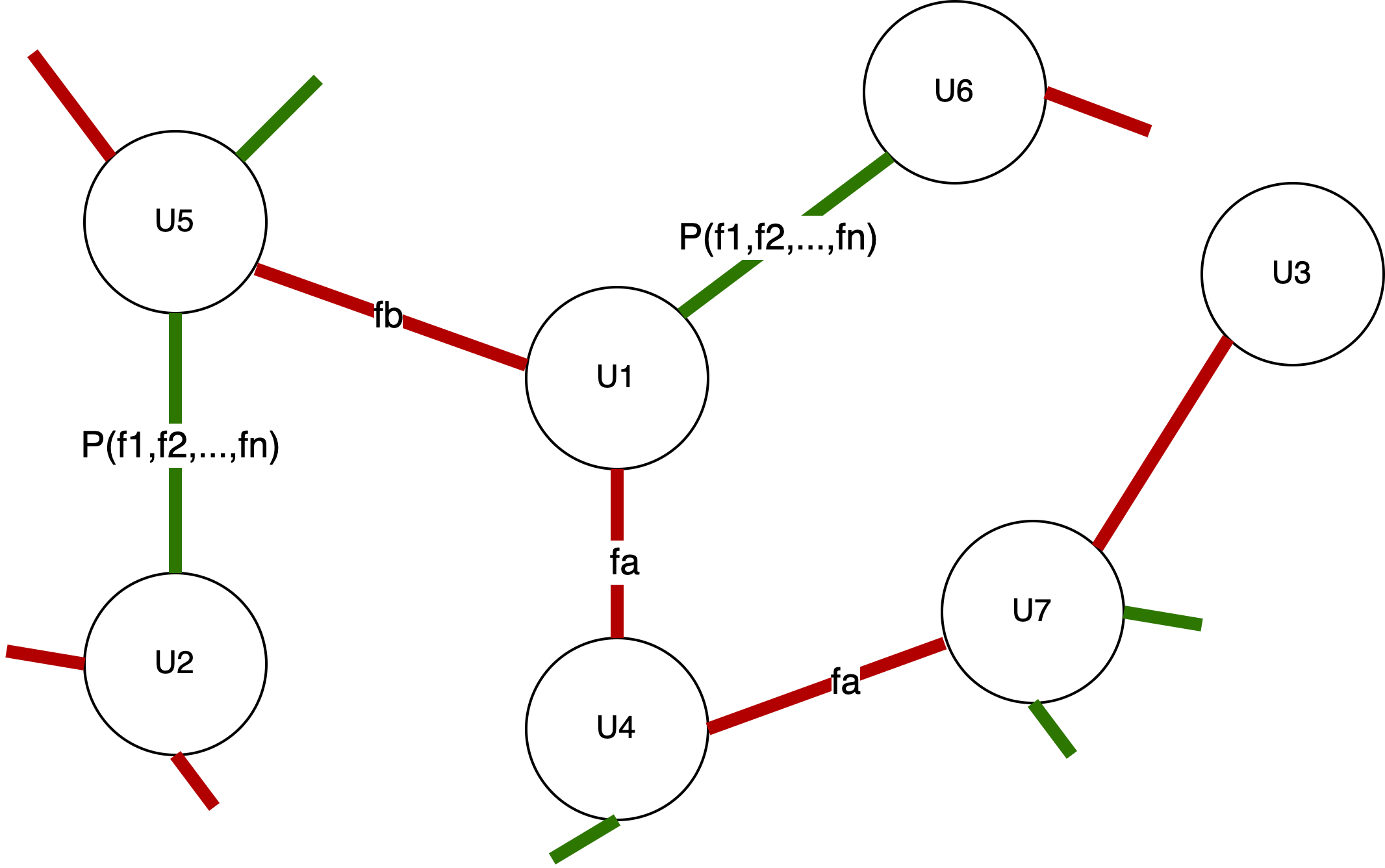}
  \caption{Graph subset showing 7 users interconnected by edges of the two types:
Heuristic edges (Red), Model based edges (Green)}
  \Description{Types of edges in our FENCE system}
  \label{fig:combined_edge_graph}
\end{figure}

Here are the terms we will use in the following algorithms

\begin{itemize}
    \item Edge(undirected) between 2 users is $E = \left( u_{1},u_{2}\right)$ where $U$ is user (Node)
    \item User attributes are defined as $F_u = \left\{ a_1, a_2, a_3, \cdots, a_p \right\}$
    \item Edge Features are defined as $F_e = \left\{ f_1, f_2, f_3, \cdots, f_q \right\}$  where p is the number of user attributes and q is the number of edge features. $f_i = C(a_{1_{i}},a_{2_{i}})$ where C is comparison operator to output a hash representing user association with respect to that attribute type.
\end{itemize}

After we create the edges heuristically or with model predictions, we combine the edges as seen in figure \ref{fig:combined_edge_graph}. All edges once formed are treated equally whether made by heuristic or deterministic methods. The following sections detail how these edges are created.

\subsubsection {Heuristic edge}
\label{deter_edge}
\hfill\\
Deterministic edges or heuristics give us high precision on the most obvious signals of MI associations. Even while showing a lower recall, they act as an important first line of defense. A similar approach is described here \cite{xu2021deep}. We add them to our graph as below:
\begin{enumerate}
    \item Create attribute-wise associations between new users and old users from a set of heuristically curated features $A$
    \item The resulting associations are considered the edge set $E$
\end{enumerate}
\hfill\\
Example (Edge creation): \textit{Take 2 nodes (users) and a proxy edge feature such as user IP. To create the deterministic edge using this proxy feature, we will compare the IP address of the users and create an edge if there is an equality between both values}

\subsubsection{Model based edge}
\label{prob_edge}
\hfill\\
Edge prediction acts as the second line of defense in case the first line of defense gets bypassed. While a probabilistic setup is prone to false positives, it is comparatively more adaptive to constantly changing patterns of duplicate account creation. 

As seen in algorithm \ref{alg:association}, Edge data is prepared to have the independent feature set $X$ in order to predict the probability $p$ for an edge to exist between 2 users. This enables us to formulate the link prediction as a binary class classification problem. $X$ is the independent feature set $F_e = C(F_{u_{1}},F_{u_{2}})$.  These features are computed by comparing the user attributes for exact matches, partial matches, and similarity using distance measures. Using user feature set $F_u$ of $U_1$ and $U_2$ we create edge level attributes $F_e$. 



\begin{algorithm}[h]
  \caption{Model-based edge prediction}
  \label{alg:association}
  \textbf{Input:} edge-id, curated features $F_u$ \\
  \textbf{Output:} edgePresent propensity
  
  \begin{enumerate}
    \item Using cross join between a user(new) and all users registered before them(old) create a list of all possible edges. This acts as an edge with an edge-id. Edge-id is a concatenation of new and old userid.
    \item Create edge features $F_e$ from user attributes $F_u$ of new and old users for all the edges
    \item Predict edgePresent propensity $P(\text{edge}) = f(F_e)$
    \item Consider edges with $P(\text{edge}) > \text{threshold}$ as final edges
  \end{enumerate}
\end{algorithm}

To get $f$, we used a distributed training flavor of the Random Forest model from the PySpark ML library \cite{sayed2018predicting}. The data for training is generated as follows:
\begin{itemize}
    \item Positive (MI edge present) edges are sampled from manually validated edges from historical runs
    \item Negative edges are sampled from a cross-join between validated unique (non-MI) users 
\end{itemize}

\subsection{Graph creation:}

\begin{figure}[h]
  \centering
  \includegraphics[width=\linewidth]{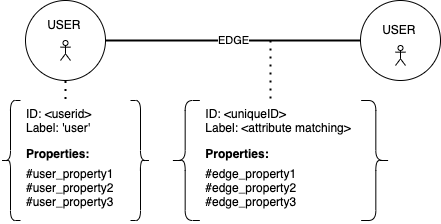}
  \caption{Data model for our graph representation}
  \Description{Data model for graph}
  \label{fig:data_model_graph}
\end{figure}

Our user graph \(G = \left ( E, V \right )\) is a structure consisting of a set of vertices V which represent the users of Dream11, and a set of edges E which represent the association between the vertices. An edge \(e \in  E\) is denoted in the form \(e = \left\{ x,y\right\}\), where the vertices \(x,y \in V\). Two vertices or users are connected by the edge \(e = \left\{ x,y\right\}\) if there exists an edge between x and y. The data model is detailed in figure \ref{fig:data_model_graph}.
In our use case, we utilize \textit{AWS Neptune}, a managed and scalable graph database. It is compatible with \textit{Gremlin Query Language}, which enables us to retrieve and modify data within the graph. 
The heuristic and model-based edges mentioned in section 5.3 are dynamically added to this graph at different time intervals. This is how our graph stays updated.

\subsection{Multiple ID Account Detection}

Now we are at a stage where we have all users in a Graph with edges related to FPV associations between them. Notably, MI accounts exhibit dense connections amongst each other and are generally found in clusters, while benign accounts have sparse connections. The identification of MI accounts entails the detection of densely connected components or subgraphs within the registration graph, which are referred to as user clusters. Each cluster is assigned an individual score, and registration accounts belonging to clusters with scores surpassing a certain threshold are categorized as FPV accounts. The cluster scoring function is a statistical algorithm that takes into consideration various factors, including cluster size, node types within the cluster, the presence of n-degree connections, and scenarios involving multiple accounts utilized by family members on the same device.
We will delve into the different approaches, their associated challenges, and the strategies employed to overcome these challenges.
\subsubsection{\textbf{Approach 1: Connected Component over Neptune using Gremlin}}
\label{sec:app1}
Once we have our user added to our registration graph we can use graph query language to retrieve all the users connected to incoming new user registration via n-degree connections using graph query languages like Gremlin,  The time complexity of searching all the connected components of a graph depends on various factors, including the size of the graph, the structure of the graph.\\
\textbf{Limitation }: Our investigation revealed that distributed graph databases like AWS Neptune are not optimized for the specific queries needed to retrieve connected components in real time for highlighting FPV users. As the number of edges and vertices increases, the latency also escalates \cite{neptuneBenchmark}, rendering it unsuitable for real-time detection purposes.

\subsubsection{\textbf {Approach 2: Distributed Connected Components}}
\label{sec:con_com}
The Alternating algorithm of "Connected Components in MapReduce and Beyond" \cite{connectedComp} can be used which aims to fetch the connected components in linear time using a distributed approach. The algorithm has 2 core operations namely Small Star and Large Star performed in alternating sequences until convergence, to fetch connected components.

\begin{itemize}
\item \textbf{Large star}

A large star operation, when applied to a node in a sub-graph, connects the minimum node in the sub-graph to all the nodes that are larger than the node itself. The large star operation is illustrated in the figure below for node 1. The large star connects 1 to all nodes with labels strictly larger than 8. In particular, 1 is not connected to 8.

\begin{figure}[H]
  \centering
  \includegraphics[width=0.9\linewidth]{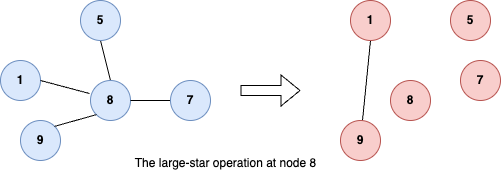}
  \caption{Large Star Operation}
  \label{fig:large-star}
\end{figure}

\item \textbf{Small star}

A small star operation, when applied to a node in a sub-graph, connects the minimum node in the sub-graph to all the nodes that are equal to or smaller than the node itself. The small star operation is illustrated in the figure below for node 1. The small star connects 1 to all node labels no larger than 8. In particular, 1 is connected to 8.

\begin{figure}[H]
  \centering
  \includegraphics[width=0.9\linewidth]{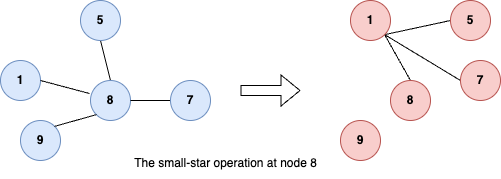}
  \caption{Small Star Operation}
  \label{fig:small-star}
\end{figure}
\end{itemize}

\begin{algorithm}
\caption{Alternating Algorithm}
\begin{algorithmic}
\STATE \textbf{Input}: Edges ($u,$v) as set of key-value pairs <$u;$v> 
\STATE \textbf{Input}: A unique label $l_v$ for every node $v \in A$.
\REPEAT
\STATE \textbf{large-star}
\STATE \textbf{small-star}
\UNTIL Convergence
\end{algorithmic}
\end{algorithm}

Within the framework of the Alternating Algorithm, "convergence" denotes the point at which no additional modifications occur in the number of nodes and vertices, signifying the attainment of the final state where all connected components have been successfully identified. The algorithm achieves convergence with a time complexity of $O(n)$. For distributed execution, we employ the map-reduce functionality provided by Apache Spark. However, this paper does not delve into the intricacies of the algorithm's proofs and correctness; interested readers can find further details and proofs in \cite{connectedComp}.\\
\textbf{Limitation }: While approach 2 (refer to section \ref{sec:con_com}) aims to address the limitations of approach 1 (refer to section \ref{sec:app1}) by achieving linear time complexity to fetch all the connected-components, it still necessitates the computation of all combinations of edges and vertices to determine every connected component in the graph. This aspect also poses challenges for our real-time detection approach.

\subsubsection{\textbf {Approach 3: Caching-enhanced Multiple-Id detection}}
\label{'app3'}
Approach 3 is a hybrid solution that combines the use of a Graph DB and Distributed Connected Components, along with a cache (in this case, Redis). This approach overcomes the limitations of the previously discussed approaches and allows for real-time detection of FPV users. The main idea behind this approach is that as the size of a cluster or community increases, or in the case of densely connected components, any user who forms an edge with a user from an existing cluster is a potential FPV user. To detect registered users in real-time, the following steps are involved:
\begin{itemize}
\item\textbf{Step 1}: The process begins by running a connected component algorithm on all registered users to determine the clusters or communities they belong to.

\item\textbf{Step 2}: Once the user-cluster mapping is obtained, this information is stored in a key-value-based cache for efficient lookup with $O(1)$ complexity.

\item\textbf{Step 3}: In this step, a Graph DB is used to fetch the direct connections (1-degree connections) of the newly registered user. Fetching the 1-degree connections of a node can be performed with $O(1)$ time complexity, which is faster compared to fetching the entire connected components.

\item\textbf{Step 4}: Once all the user nodes connected via 1-degree connections to the newly registered node are obtained, the cluster is fetched from the cache, and the new user is assigned to the same cluster. It is worth noting that if a user falls into two or more clusters, they are attached to the cluster with the most nodes, and clusters are reconciled in a separate reconciliation job.
\end{itemize}

\textbf{Limitation }: Overall, this hybrid approach combines the benefits of a Graph DB, Distributed Connected Components, and a cache to efficiently detect FPV users in real time while addressing the limitations of previous approaches however, the approach might fail to highlight certain users due to outdated cache where the user creates an edge among several smaller cluster where the score does not breach the detection threshold thereby reducing re-call number slightly. As we use Real-time FENCE as the first line of defense in our overall fraud detection ecosystem, These missed users are caught in Batch Flow.

Overall, this hybrid approach combines the benefits of a Graph DB, Distributed Connected Components, and a cache to efficiently detect FPV users in real time while addressing the limitations of previous approaches.

\subsection{User Actioning }
All the scoring is done on a cluster (connected component) level and users whose cluster score is above a certain threshold are automatically blocked as conclusive multiple IDs of a single user. For any other users captured by the system, there is a manual validation check before taking any action.

Owing to the sensitive nature of the problem, we depend on human intervention at various stages of model development as well as validation. So this system is a classic application of \textbf{Human-in-the-loop (HITL)} machine learning.
\begin{itemize}
    \item \textit{Changing patterns} - Continuous real-time actioning leads to fast-evolving user patterns. To monitor and keep up with the same, we share stratified random samples of users with the risk operations team who validate these manually and report any new patterns that our models might have missed. 
    \item \textit{Validation and Feedback} - Multiple ID clusters with lower scores are processed through manual flows. Additionally, any new version of the deployed models is also passed through the experimental pipeline which is then screened by the risk operations team. The feedback from these validations is utilized for future training. 

\end{itemize}

\subsection{Maintenance and monitoring}
MLFlow has been used end to end for training, inferencing, and maintaining this system.
\subsubsection{Cluster-Reconciliation}
Approach 3 (\ref{'app3'}) takes advantage of a user-cluster mapping built by executing connected components on existing edges. When a new user registers, this mapping is saved and cached, allowing for a speedy look-up. However, if a new user registration results in the connectivity of previously disconnected clusters, the cache must be updated. To do this, we execute a reconciliation operation that identifies and accounts for all cluster mergers, ensuring that the cache reflects the most recent state.
\subsubsection{Training}
\hfill\\
Spark distributed set-up has been used to train models and each run was logged on MLFlow for comparison and benchmarking. The models are logged as artifacts and registered to the MLFlow registry from where they are referenced at the time of inference.
\subsubsection{Inference}
\hfill\\
Both, our real-time and non-real-time applications reference the MLFlow model registry to inference on all incoming records.
\subsubsection{Maintenance}
\hfill\\
All versions of the model are logged and maintained on MLFlow. After an upgrade to the model, its new version is registered as a “staging” model. This model is tested in a parallel setup before being promoted to “production”.
\subsubsection{Model Monitoring}
\hfill\\
To monitor the performance of the offline and online models we send a certain \% of highlights for manual reviews even from all flows. This \% varies based on the user registration scale on the platform but generally ranges between 1\% to 10\%. This helps us monitor if there is any significant drift in our models in terms of the False Positive Rate and Precision.

\subsubsection{User Feedback}
\hfill\\
Any feedback on the action taken by us on the users is logged in the system for model retraining.

\section{Metrics}

\subsection{Approach Comparison benchmarking}

\begin{figure}[H]
    \centering
    \includegraphics[width=\linewidth]{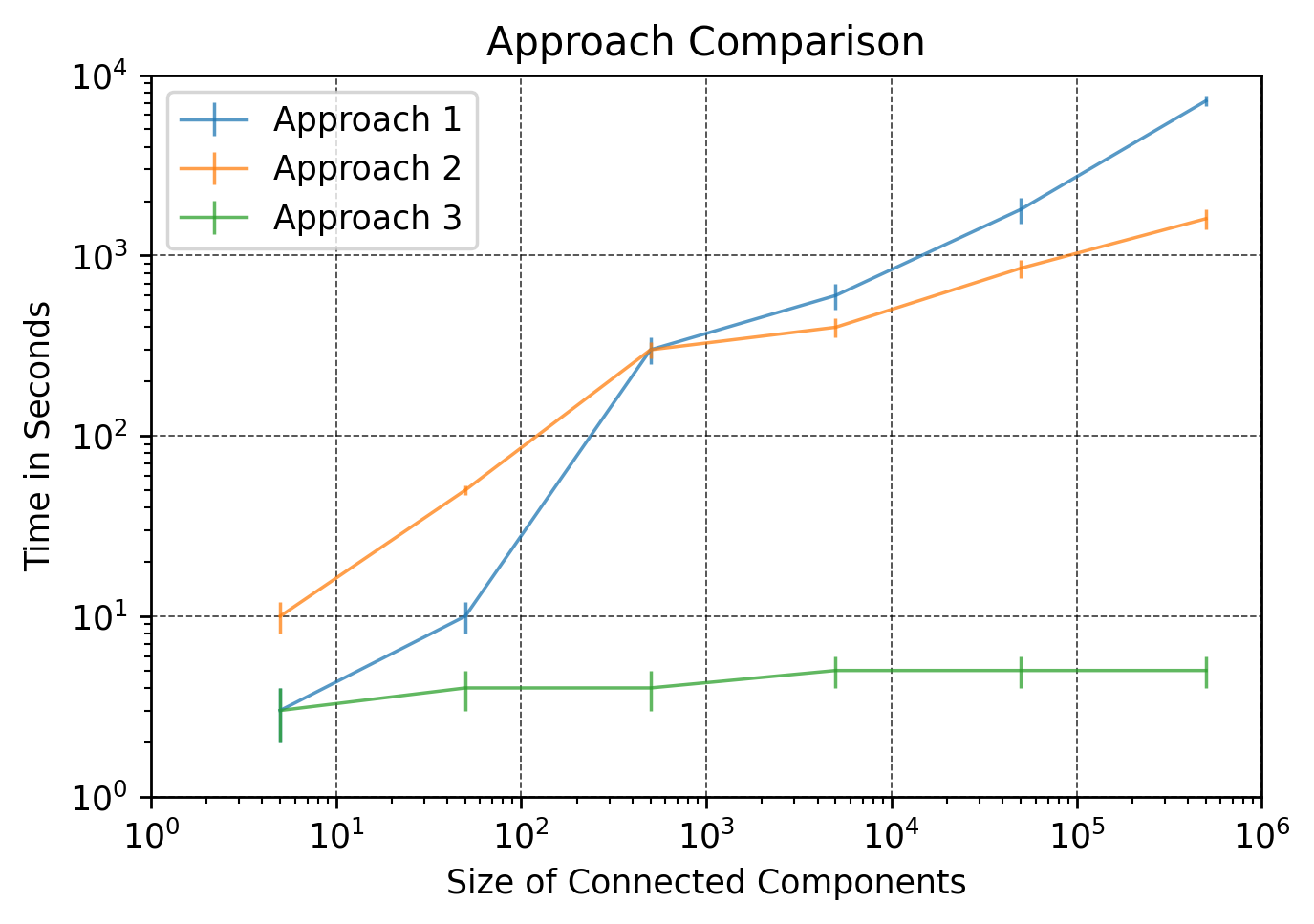}
    \caption{Runtime Comparison of aforementioned approaches as the size of connected components increases}
    \label{fig:approach_cmp}
\end{figure}

Our Real-time production pipeline built on Approach 3(section \ref{'app3'}) processes users within \textbf{4 seconds} of their registration. This remains roughly constant irrespective of the size of the connected component as depicted in figure \ref{fig:approach_cmp}.

\subsection{Edge scale analysis}

\begin{figure}[H]
    \centering
    \includegraphics[width=\linewidth]{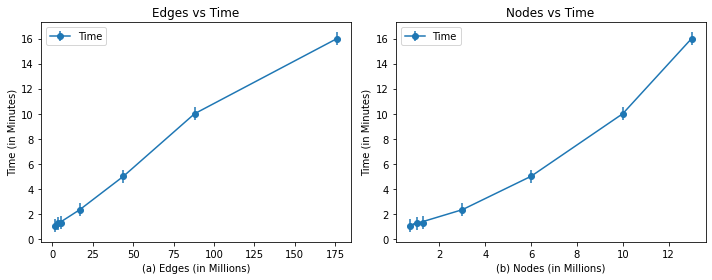}
    \caption{Left fig (a): Edge scale comparison, Right fig (b): Node Scale comparison}
    \label{fig:cc}
\end{figure}

Figure \ref{fig:cc} shows benchmarking of the run-times of Alternating Algorithm for different Edge and Node Counts on AWS instance (6 $*$  r5d.4xlarge).

\subsection{Business Impact}

Post deployment of this system we observed a relative decrease of \textbf{86\%} in system FPV. This brought down the overall system FPV to an acceptable level and has been sustained at the same level since. 

\subsection{Model Performance}

\begin{table}[h]
\centering
\begin{tabular}{|c|c|c|}
\hline
Flow  & Online Precision & Training Recall \\
\hline
Automated Flow & 96.7\% & 55.2\% \\
Manual Flow & 70.2\% & 86.4\% \\
\hline
\end{tabular}
\caption{Online precision and training recall for automated and manual flows}
\label{tab:mytable}
\end{table}
Please note the training recall numbers are based on the known ground truth available at the time of training. Online precision is calculated on active detection based on user and operations team's feedback for automated and manual flows respectively. The automated flow has high precision and low recall due to higher thresholds. This high standard is maintained so ensure lower false positives in automated flows so that genuine users are not impacted. Since the manual flow has a human in the loop the threshold is lower. This leads to improved recall with lower precision of the model.

\section{Conclusion and Future Scope}

In this research paper, we have proposed a solution to address the issue of detecting multiple-ID accounts in fantasy gaming platforms. Our approach combines supervised and unsupervised methods, utilizing a graph database to facilitate the detection process. We have introduced both heuristical and model-based techniques to enhance the graph, and we have conducted a bench-marking study comparing three different approaches for identifying connected components of multiple ID users in real-time. Our results indicate that caching enhanced community detection is the most effective method in a real-time setup.

Our current work opens up several avenues for further exploration. Firstly, our existing system assigns the same score to all members within a connected component. To achieve more granularity, we plan to incorporate label propagation, similar to the message passing technique in Graph Convolutional Networks, to assign user-specific scores.

Secondly, we have a number of new features in the pipeline. For instance, we are aware that standalone device attributes can be easily manipulated by smart FPV users. To address this, we are currently experimenting with the ability to combine device-related attributes and compute a single unique device fingerprint for each user device. This fingerprint can then be added as an edge to the graph, contributing to a more comprehensive detection process.

Furthermore, approximately 50\% of our highlighted cases currently rely on manual flows. Our objective is to gain more confidence across all FPV highlights and transition 80\% of the processing to automated flows. This will streamline the detection process and enhance overall efficiency.


\begin{acks}
To Data Engineering for helping us set up the graph DB, Risk Operation for all ground-truth evaluations, and Analytics for valuable insights.

\end{acks}

\bibliographystyle{ACM-Reference-Format}
\bibliography{fence-refs}


\begin{thebibliography}{14}


\ifx \showCODEN    \undefined \def \showCODEN     #1{\unskip}     \fi
\ifx \showDOI      \undefined \def \showDOI       #1{#1}\fi
\ifx \showISBNx    \undefined \def \showISBNx     #1{\unskip}     \fi
\ifx \showISBNxiii \undefined \def \showISBNxiii  #1{\unskip}     \fi
\ifx \showISSN     \undefined \def \showISSN      #1{\unskip}     \fi
\ifx \showLCCN     \undefined \def \showLCCN      #1{\unskip}     \fi
\ifx \shownote     \undefined \def \shownote      #1{#1}          \fi
\ifx \showarticletitle \undefined \def \showarticletitle #1{#1}   \fi
\ifx \showURL      \undefined \def \showURL       {\relax}        \fi
\providecommand\bibfield[2]{#2}
\providecommand\bibinfo[2]{#2}
\providecommand\natexlab[1]{#1}
\providecommand\showeprint[2][]{arXiv:#2}

\bibitem[Atemezing(2021)]%
        {neptuneBenchmark}
\bibfield{author}{\bibinfo{person}{Ghislain~Auguste Atemezing}.} \bibinfo{year}{2021}\natexlab{}.
\newblock \showarticletitle{Empirical Evaluation of a Cloud-Based Graph Database: the Case of Neptune}. In \bibinfo{booktitle}{\emph{Knowledge Graphs and Semantic Web: Third Iberoamerican Conference and Second Indo-American Conference, KGSWC 2021, Kingsville, Texas, USA, November 22--24, 2021, Proceedings 3}}. Springer, \bibinfo{pages}{31--46}.
\newblock


\bibitem[Boshmaf et~al\mbox{.}(2016)]%
        {boshmaf2016integro}
\bibfield{author}{\bibinfo{person}{Yazan Boshmaf}, \bibinfo{person}{Dionysios Logothetis}, \bibinfo{person}{Georgos Siganos}, \bibinfo{person}{Jorge Ler{\'\i}a}, \bibinfo{person}{Jose Lorenzo}, \bibinfo{person}{Matei Ripeanu}, \bibinfo{person}{Konstantin Beznosov}, {and} \bibinfo{person}{Hassan Halawa}.} \bibinfo{year}{2016}\natexlab{}.
\newblock \showarticletitle{{\'I}ntegro: Leveraging victim prediction for robust fake account detection in large scale OSNs}.
\newblock \bibinfo{journal}{\emph{Computers \& Security}}  \bibinfo{volume}{61} (\bibinfo{year}{2016}), \bibinfo{pages}{142--168}.
\newblock


\bibitem[Cao et~al\mbox{.}(2014)]%
        {cao2014uncovering}
\bibfield{author}{\bibinfo{person}{Qiang Cao}, \bibinfo{person}{Xiaowei Yang}, \bibinfo{person}{Jieqi Yu}, {and} \bibinfo{person}{Christopher Palow}.} \bibinfo{year}{2014}\natexlab{}.
\newblock \showarticletitle{Uncovering large groups of active malicious accounts in online social networks}. In \bibinfo{booktitle}{\emph{Proceedings of the 2014 ACM SIGSAC conference on computer and communications security}}. \bibinfo{pages}{477--488}.
\newblock


\bibitem[Danezis and Mittal(2009)]%
        {danezis2009sybilinfer}
\bibfield{author}{\bibinfo{person}{George Danezis} {and} \bibinfo{person}{Prateek Mittal}.} \bibinfo{year}{2009}\natexlab{}.
\newblock \showarticletitle{Sybilinfer: Detecting sybil nodes using social networks.}. In \bibinfo{booktitle}{\emph{Ndss}}. San Diego, CA, \bibinfo{pages}{1--15}.
\newblock


\bibitem[Egele et~al\mbox{.}(2015)]%
        {egele2015towards}
\bibfield{author}{\bibinfo{person}{Manuel Egele}, \bibinfo{person}{Gianluca Stringhini}, \bibinfo{person}{Christopher Kruegel}, {and} \bibinfo{person}{Giovanni Vigna}.} \bibinfo{year}{2015}\natexlab{}.
\newblock \showarticletitle{Towards detecting compromised accounts on social networks}.
\newblock \bibinfo{journal}{\emph{IEEE Transactions on Dependable and Secure Computing}} \bibinfo{volume}{14}, \bibinfo{number}{4} (\bibinfo{year}{2015}), \bibinfo{pages}{447--460}.
\newblock


\bibitem[Kiveris et~al\mbox{.}(2014)]%
        {connectedComp}
\bibfield{author}{\bibinfo{person}{Raimondas Kiveris}, \bibinfo{person}{Silvio Lattanzi}, \bibinfo{person}{Vahab Mirrokni}, \bibinfo{person}{Vibhor Rastogi}, {and} \bibinfo{person}{Sergei Vassilvitskii}.} \bibinfo{year}{2014}\natexlab{}.
\newblock \showarticletitle{Connected Components in MapReduce and Beyond}.
\newblock \bibinfo{journal}{\emph{Proceedings of the 5th ACM Symposium on Cloud Computing, SOCC 2014}}, \bibinfo{pages}{1--13}.
\newblock
\urldef\tempurl%
\url{https://doi.org/10.1145/2670979.2670997}
\showDOI{\tempurl}


\bibitem[Lazemi and Ebrahimpour-Komleh(2014)]%
        {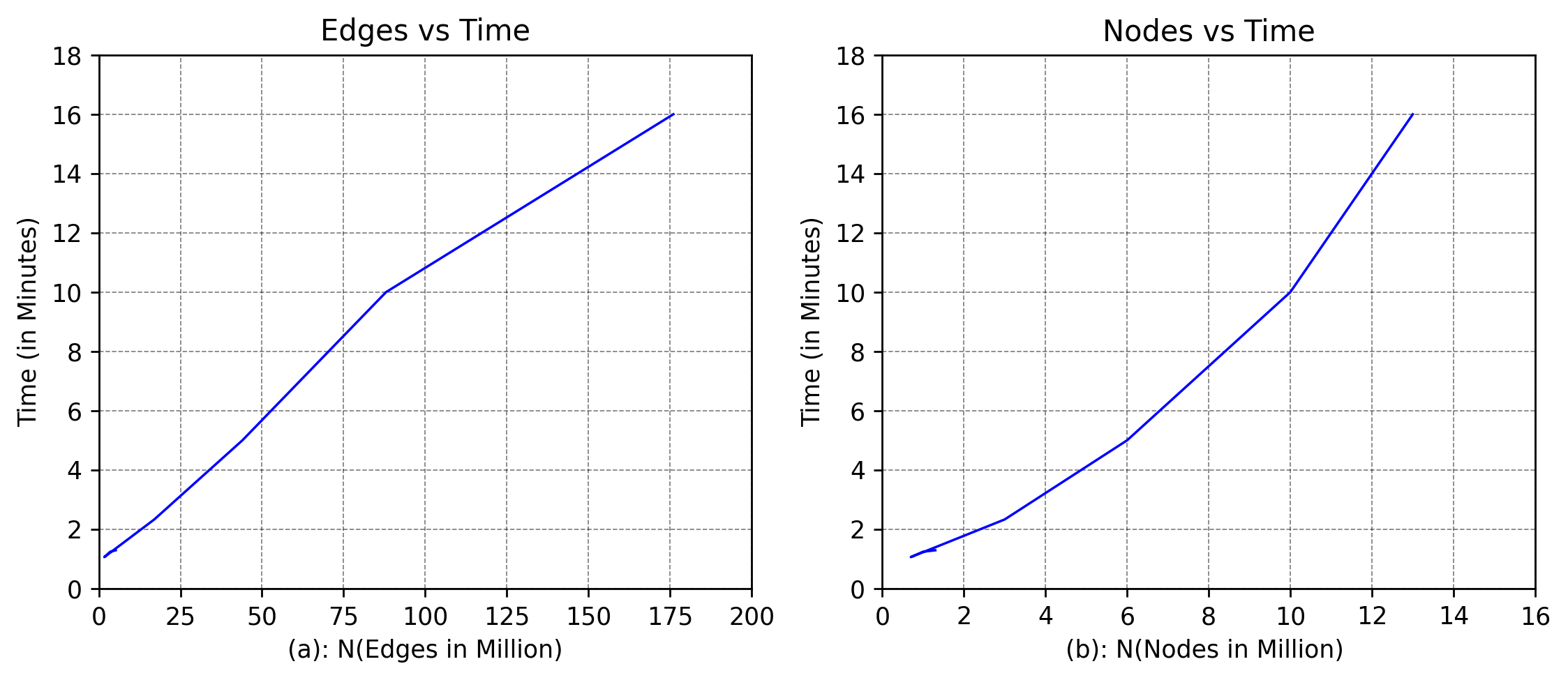}
\bibfield{author}{\bibinfo{person}{Soghra Lazemi} {and} \bibinfo{person}{Hossein Ebrahimpour-Komleh}.} \bibinfo{year}{2014}\natexlab{}.
\newblock \showarticletitle{Computing connected components of graphs}.
\newblock \bibinfo{journal}{\emph{International Journal of Applied Mathematical Research}}  \bibinfo{volume}{3} (\bibinfo{date}{09} \bibinfo{year}{2014}), \bibinfo{pages}{508}.
\newblock
\urldef\tempurl%
\url{https://doi.org/10.14419/ijamr.v3i4.3189}
\showDOI{\tempurl}


\bibitem[Liang et~al\mbox{.}(2021)]%
        {liang2021unveiling}
\bibfield{author}{\bibinfo{person}{Xiao Liang}, \bibinfo{person}{Zheng Yang}, \bibinfo{person}{Binghui Wang}, \bibinfo{person}{Shaofeng Hu}, \bibinfo{person}{Zijie Yang}, \bibinfo{person}{Dong Yuan}, \bibinfo{person}{Neil~Zhenqiang Gong}, \bibinfo{person}{Qi Li}, {and} \bibinfo{person}{Fang He}.} \bibinfo{year}{2021}\natexlab{}.
\newblock \showarticletitle{Unveiling fake accounts at the time of registration: An unsupervised approach}. In \bibinfo{booktitle}{\emph{Proceedings of the 27th ACM SIGKDD Conference on Knowledge Discovery \& Data Mining}}. \bibinfo{pages}{3240--3250}.
\newblock


\bibitem[Mondal et~al\mbox{.}(2012)]%
        {mondal2012defending}
\bibfield{author}{\bibinfo{person}{Mainack Mondal}, \bibinfo{person}{Bimal Viswanath}, \bibinfo{person}{Allen Clement}, \bibinfo{person}{Peter Druschel}, \bibinfo{person}{Krishna~P Gummadi}, \bibinfo{person}{Alan Mislove}, {and} \bibinfo{person}{Ansley Post}.} \bibinfo{year}{2012}\natexlab{}.
\newblock \showarticletitle{Defending against large-scale crawls in online social networks}. In \bibinfo{booktitle}{\emph{Proceedings of the 8th international conference on Emerging networking experiments and technologies}}. \bibinfo{pages}{325--336}.
\newblock


\bibitem[Sadowski and Rathle(2014)]%
        {sadowski2014fraud}
\bibfield{author}{\bibinfo{person}{Gorka Sadowski} {and} \bibinfo{person}{Philip Rathle}.} \bibinfo{year}{2014}\natexlab{}.
\newblock \showarticletitle{Fraud detection: Discovering connections with graph databases}.
\newblock \bibinfo{journal}{\emph{White Paper-Neo Technology-Graphs are Everywhere}}  \bibinfo{volume}{13} (\bibinfo{year}{2014}).
\newblock


\bibitem[Sayed et~al\mbox{.}(2018)]%
        {sayed2018predicting}
\bibfield{author}{\bibinfo{person}{Hend Sayed}, \bibinfo{person}{Manal~A Abdel-Fattah}, {and} \bibinfo{person}{Sherif Kholief}.} \bibinfo{year}{2018}\natexlab{}.
\newblock \showarticletitle{Predicting potential banking customer churn using apache spark ML and MLlib packages: a comparative study}.
\newblock \bibinfo{journal}{\emph{International Journal of Advanced Computer Science and Applications}} \bibinfo{volume}{9}, \bibinfo{number}{11} (\bibinfo{year}{2018}).
\newblock


\bibitem[Times(2023)]%
        {amit_interview}
\bibfield{author}{\bibinfo{person}{Economic Times}.} \bibinfo{year}{2023}\natexlab{}.
\newblock \bibinfo{booktitle}{\emph{Technology will bridge gap between sports and sports fans: Dream11 CTO Amit Sharma}}.
\newblock
\urldef\tempurl%
\url{https://telecom.economictimes.indiatimes.com/news/internet/technology-will-bridge-gap-between-sports-and-sports-fans-dream-sports-cto-amit-sharma/99430836}
\showURL{%
Retrieved April 13, 2023 from \tempurl}


\bibitem[Wang et~al\mbox{.}(2017)]%
        {wang2017sybilscar}
\bibfield{author}{\bibinfo{person}{Binghui Wang}, \bibinfo{person}{Le Zhang}, {and} \bibinfo{person}{Neil~Zhenqiang Gong}.} \bibinfo{year}{2017}\natexlab{}.
\newblock \showarticletitle{SybilSCAR: Sybil detection in online social networks via local rule based propagation}. In \bibinfo{booktitle}{\emph{IEEE INFOCOM 2017-IEEE Conference on Computer Communications}}. IEEE, \bibinfo{pages}{1--9}.
\newblock


\bibitem[Xu et~al\mbox{.}(2021)]%
        {xu2021deep}
\bibfield{author}{\bibinfo{person}{Teng Xu}, \bibinfo{person}{Gerard Goossen}, \bibinfo{person}{Huseyin~Kerem Cevahir}, \bibinfo{person}{Sara Khodeir}, \bibinfo{person}{Yingyezhe Jin}, \bibinfo{person}{Frank Li}, \bibinfo{person}{Shawn Shan}, \bibinfo{person}{Sagar Patel}, \bibinfo{person}{David Freeman}, {and} \bibinfo{person}{Paul Pearce}.} \bibinfo{year}{2021}\natexlab{}.
\newblock \showarticletitle{Deep entity classification: Abusive account detection for online social networks}. In \bibinfo{booktitle}{\emph{30th $\{$USENIX$\}$ Security Symposium ($\{$USENIX$\}$ Security 21)}}.
\newblock


\end{thebibliography}

\appendix

\end{document}